\documentclass[conference]{IEEEtran}
\IEEEoverridecommandlockouts
\usepackage{cite}
\usepackage{amsmath,amssymb,amsfonts}
\usepackage{algorithmic}
\usepackage{graphicx}
\usepackage{hyperref} 
\usepackage{pifont}
\usepackage{booktabs}
\usepackage{textcomp}
\usepackage{xcolor}
\newcommand{\RNum}[1]{\uppercase\expandafter{\romannumeral #1\relax}}

\def\BibTeX{{\rm B\kern-.05em{\sc i\kern-.025em b}\kern-.08em
    T\kern-.1667em\lower.7ex\hbox{E}\kern-.125emX}}
\begin{document}

\title{TAMISeg: Text-Aligned Multi-scale Medical Image Segmentation with Semantic Encoder Distillation\\
\thanks{This work was supported in part by the Natural Science Foundation of Chongqing (CSTB2024NSCQ-KJFZZDX0036); Central University basic research young teachers and students research ability promotion sub-project(2023CDJYGRH-ZD06); Chongqing Technology Innovation and Application Development Project (CSTB2023TIAD-KPX0050); a key joint project of Chongqing Health Commission and Science and Technology Bureau (2024ZDXM007)
\\
* indicates the corresponding authors.}
}

\author{\IEEEauthorblockN{1\textsuperscript{st} Qiang Gao}
\IEEEauthorblockA{\textit{Department of Data Science and AI} \\
\textit{Monash University}\\
Melbourne, Australia\\
Qiang.Gao@monash.edu}
\and
\IEEEauthorblockN{2\textsuperscript{nd} Yi Wang*}
\IEEEauthorblockA{\textit{College of Computer Science} \\
\textit{Chongqing University}\\
Chongqing, China \\
yiwang@cqu.edu.cn}
\and
\IEEEauthorblockN{3\textsuperscript{rd} Yong Zhang}
\IEEEauthorblockA{\textit{College of Computer Science} \\
\textit{Chongqing University}\\
Chongqing, China \\
zhangyong7630@cqu.edu.cn}
\and
\IEEEauthorblockN{4\textsuperscript{th} Yong Li}
\IEEEauthorblockA{\textit{College of Computer Science} \\
\textit{Chongqing University}\\
Chongqing, China \\
yongli@cqu.edu.cn}
\and
\IEEEauthorblockN{4\textsuperscript{th} Yongbing Deng}
\IEEEauthorblockA{\textit{Chongqing University Central Hospital} \\
\textit{School of Medicine}\\
Chongqing, China \\
dyb0913@cqu.edu.cn}
\and
\IEEEauthorblockN{5\textsuperscript{th} Lan Du}
\IEEEauthorblockA{\textit{Department of Data Science and AI} \\
\textit{Monash University}\\
Melbourne, Australia \\
Lan.Du@monash.edu}
\and
\IEEEauthorblockN{6\textsuperscript{th} Cunjian Chen*}
\IEEEauthorblockA{\textit{Department of Data Science and AI} \\
\textit{Monash University}\\
Melbourne, Australia \\
Cunjian.Chen@monash.edu}
}

\author{
\IEEEauthorblockN{Qiang Gao$^{1}$, Yi Wang$^{2*}$, Yong Zhang$^{2}$, Yong Li$^{2}$, Yongbing Deng$^{3}$, Lan Du$^{1}$, Cunjian Chen$^{1*}$}
\IEEEauthorblockA{$^{1}$Department of Data Science and AI, Monash University, Melbourne, Australia}
\IEEEauthorblockA{$^{2}$College of Computer Science, Chongqing University, Chongqing, China}
\IEEEauthorblockA{$^{3}$Chongqing University Central Hospital, School of Medicine, Chongqing, China}
\IEEEauthorblockA{Qiang.Gao@monash.edu, yiwang@cqu.edu.cn, zhangyong7630@cqu.edu.cn, yongli@cqu.edu.cn,}
\IEEEauthorblockA{dyb0913@cqu.edu.cn, Lan.Du@monash.edu, Cunjian.Chen@monash.edu}
}

\maketitle

\begin{abstract}
Medical image segmentation remains challenging due to limited fine-grained annotations, complex anatomical structures, and image degradation from noise, low contrast, or illumination variation. We propose TAMISeg, a text-guided segmentation framework that incorporates clinical language prompts and semantic distillation as auxiliary semantic cues to enhance visual understanding and reduce reliance on pixel-level fine-grained annotations. TAMISeg integrates three core components: a consistency-aware encoder pretrained with strong perturbations for robust feature extraction, a semantic encoder distillation module with supervision from a frozen DINOv3 teacher to enhance semantic discriminability, and a scale-adaptive decoder that segments anatomical structures across different spatial scales. Experiments on the Kvasir-SEG, MosMedData+, and QaTa-COV19 datasets demonstrate that TAMISeg consistently outperforms existing uni-modal and multi-modal methods in both qualitative and quantitative evaluations. Code will be made publicly available at \href{https://github.com/qczggaoqiang/TAMISeg}{https://github.com/qczggaoqiang/TAMISeg}.
\end{abstract}


\section{Introduction}
\label{sec:introduction}

Medical image segmentation plays an essential role in accurately delineating pathological regions, including pulmonary infectious diseases and colorectal polyps~\cite{ye2025alleviating}. It enables accurate localization of lesions, significantly supporting clinical diagnosis, therapeutic planning, and disease monitoring. Deep learning-based methods have achieved significant progress in medical image segmentation, including CNN-based architectures such as U-Net~\cite{unet} and U-Net++~\cite{zhou2018unet++}, as well as hybrid CNN-Transformer frameworks such as TransUNet~\cite{transunet}, which have been effectively applied to lesion segmentation in medical images. Despite the remarkable
progress of these methods, achieving reliable performance remains challenging due to the inherent complexity of anatomical structures, limited pixel-level details, and frequent image degradation caused by noise, low contrast, or poor illumination. 

Additionally, most existing segmentation models rely solely on visual information and require a considerable amount of pixel-level annotations to achieve satisfactory performance~\cite{transunet,zhou2018unet++,lei2025condseg}. However, obtaining such detailed annotations in clinical settings is often impractical due to the high cost and the need for significant expert involvement~\cite{jie2025diffusion}. Furthermore, visual-only approaches remain vulnerable to performance degradation under challenging conditions, such as low-contrast tissues, image noise, and uneven illumination~\cite{lei2025condseg,li2025kg}. These limitations highlight the need for more robust and semantically enriched segmentation frameworks. While CondSeg~\cite{lei2025condseg} enhances feature robustness via consistency-driven visual learning, its CNN-based design captures only local features and lacks textual semantic cues, limiting its capability in modeling global context and complex anatomical structures. Motivated by the substantial performance gains demonstrated by MedCLIP~\cite{wang2022medclip} through the integration of textual and visual modalities, the task of text-guided medical image segmentation has garnered growing interest~\cite{li2023lvit,hu2024lga,fang2025driven,ye2025alleviating,zhong2023ariadne}. In clinical practice, diagnostic text reports are typically produced in parallel with medical images, making it possible to access corresponding textual information without incurring extra annotation costs~\cite{feng2024enhancing}. However, existing text-guided segmentation methods still suffer from limited semantic discriminability, which restricts the model’s ability to extract rich contextual features from medical images. Additionally, many approaches adopt a uniform decoding mechanism, which overlooks the inherent scale variability of anatomical structures. As a result, small lesions are often missed, while large regions may be inaccurately localized. Furthermore, diffusion-based segmentation frameworks~\cite{yan2024cold,wu2024medsegdiff} have demonstrated strong performance through generative priors, yet their high computational complexity and training cost limit their efficiency and scalability in clinical applications. 

To address these limitations, we propose TAMISeg, a \textbf{t}ext-\textbf{a}ligned \textbf{m}edical \textbf{i}mage \textbf{seg}mentation framework that leverages clinical language prompts as auxiliary semantic cues to enrich visual understanding while reducing the reliance on dense pixel-level annotations. Rather than performing early fusion of text and image features, TAMISeg sequentially incorporates cross-modal knowledge through a pipeline of robust visual encoding, feature enhancement via distillation, and semantic alignment via a Cross-modal Alignment Module (CMA) with clinical text prompts. A Consistency-Aware Encoder (CAE) is pretrained with strong data perturbations to enhance robustness under noisy, low-contrast, or poorly illuminated conditions, ensuring reliable feature extraction in diverse clinical scenarios. To further improve the semantic discriminability of visual contextual representations, a Semantic Encoder Distillation (SED) module was proposed to enhance the multi-scale encoder features using supervision from a pretrained DINOv3 teacher model~\cite{simeoni2025dinov3}, which captures global semantic dependencies through its Transformer-based architecture. Although knowledge distillation has been occasionally adopted in text-guided segmentation, our framework is among the first to integrate DINOv3-based semantic encoder distillation for enhancing multi-scale visual representations prior to text alignment. In addition, TAMISeg employs a Scale-Adaptive Decoder (SAD) that accommodates anatomical structures of different sizes through specialized decoding branches, enabling precise and scale-sensitive segmentation across varied lesions.

The contributions of our work are summarized below.
\begin{itemize}

\item We propose TAMISeg, a novel text-guided medical image segmentation framework that leverages clinical language prompts as auxiliary semantic cues to reduce reliance on dense visual annotations.


\item We introduce a Semantic Encoder Distillation module, which enriches pixel-level encoder outputs via supervision from a frozen DINOv3 teacher, improving the semantic discriminability of multi-scale encoder features. 

\item We present a Scale-Adaptive Decoder that operates on text-aligned multi-scale features through parallel decoding branches specialized for anatomical structures of different sizes, improving segmentation accuracy across varying spatial scales.

\end{itemize}

\section{Methods}
\label{sec:method}

\begin{figure*}[htbp]
\centering
\setlength{\abovecaptionskip}{3pt}
\setlength{\belowcaptionskip}{-5pt}
\includegraphics[width=0.76\textwidth]{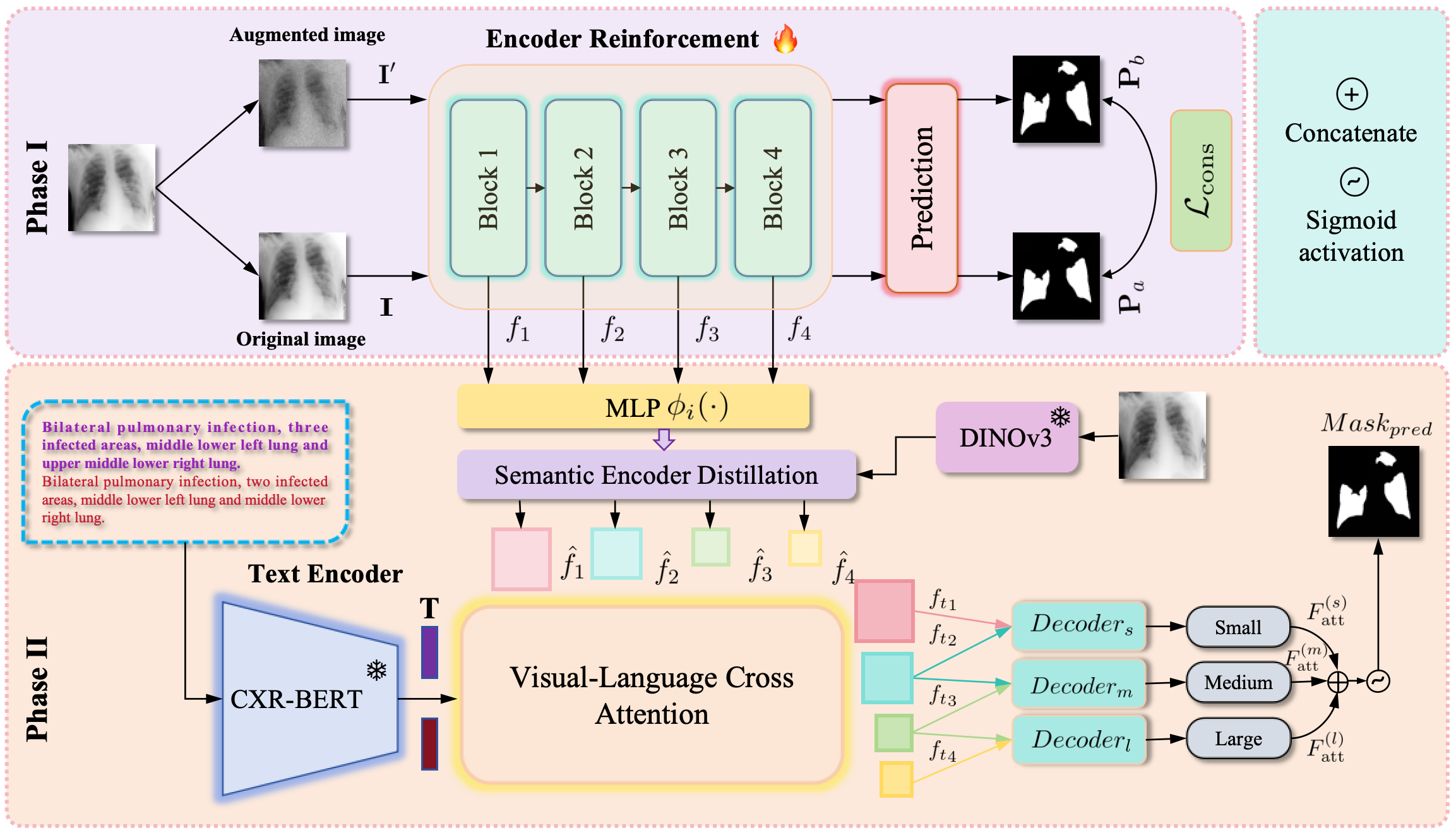}
\caption{
Overview of the proposed TAMISeg framework. \textbf{Phase~\RNum{1}}: Consistency-Aware Encoder. The encoder is trained with strong data perturbations, enforcing prediction consistency between original and augmented images to learn robust representations.
\textbf{Phase~\RNum{2}}: Cross-modal Fine-tuning. The pretrained encoder outputs four hierarchical features ($f_1$–$f_4$), which are distilled by a frozen DINOv3 teacher to enhance semantic quality. Clinical language prompts encoded by a frozen CXR-BERT model~\cite{boecking2022making} guide these features via cross-modal attention, producing aligned maps ($ {f_t}_1$–${f_t}_4$). A Scale-Adaptive Decoder (SAD) then decodes multi-scale features through parallel branches specialized for varying anatomical sizes to generate the final segmentation.}
\label{fig1}
\end{figure*}

\subsection{Overview of TAMISeg}

We propose TAMISeg, a robust text-guided framework for medical image segmentation, designed to improve visual feature quality and effectively incorporate clinical language prompts as auxiliary semantic cues. The architecture, as illustrated in Fig~\ref{fig1}, follows a two-phase design: a consistency-aware pretraining phase for high-quality feature extraction (Phase~\RNum{1}), and a fine-tuning phase where multi-scale features are enhanced via semantic encoder distillation and aligned with clinical language prompts before being decoded for precise segmentation (Phase~\RNum{2}). Specifically, the first phase begins with a Consistency-Aware Encoder, which is pretrained under strong augmentations to ensure robustness to image degradation such as noise, low contrast, and illumination shifts. This encoder generates four hierarchical feature maps ($f_1$, $f_2$, $f_3$, $f_4$), capturing visual features at multiple scales. These features are subsequently enhanced via a semantic encoder distillation process in the second phase, where a frozen DINOv3 model serves as the teacher network. The encoder's outputs are guided to match the rich semantic representations of DINOv3, yielding enhanced features ($\hat{f}_1$, $\hat{f}_2$, $\hat{f}_3$, $\hat{f}_4$) with improved discriminability. Next, these enhanced features are aligned with clinical language prompts embedded by a frozen CXR-BERT model~\cite{boecking2022making}. Through cross-modal attention, textual embeddings inject clinical semantic cues into the visual hierarchy, yielding the aligned feature maps ($ {f_t}_1 $, $ {f_t}_2 $, $ {f_t}_3 $, $ {f_t}_4 $). Finally, these multi-scale, text-aligned features are decoded by a Scale-Adaptive Decoder (SAD), which is designed to better distinguish anatomical entities of different sizes and generate tailored predictions for each scale. Each branch fuses adjacent feature levels and progressively reconstructs partial masks, which are then integrated into the final segmentation output. This unified design allows TAMISeg to achieve robust, fine-grained segmentation across diverse clinical conditions with minimal annotation dependence.

\subsection{Consistency-Aware Encoder}

To improve the robustness of feature extraction under weak illumination, low-contrast, and blurred imaging conditions, we adopt a consistency-aware pretraining strategy for the encoder of TAMISeg, implemented with a ResNet-50 backbone~\cite{lei2025condseg}. In this pretraining stage, denoted as $\mathcal{N}_{pre}$, the encoder is trained independently with a lightweight prediction head, ensuring that feature robustness is learned without assistance from external modules. Given an input image $\mathbf{I}$, a perturbed variant $\mathbf{I}' = \mathcal{T}(\mathbf{I})$ is generated through strong augmentations, including random adjustments of brightness, contrast, saturation, and hue, random conversion to grayscale, and Gaussian blurring, to simulate clinical variations such as low contrast, weak illumination, and imaging noise~\cite{lei2025condseg}. Both images are passed through $\mathcal{N}_{pre}$ to produce predictions $\mathbf{P}_a = \mathcal{N}_{pre}(\mathbf{I})$ and $\mathbf{P}_b = \mathcal{N}_{pre}(\mathbf{I}')$. Each prediction is supervised against the ground truth $\mathbf{G}$ using a hybrid loss that combines binary cross-entropy and Dice losses, constraining both $\mathbf{P}_a$ and $\mathbf{P}_b$: 
\begin{equation}
\mathcal{L}_{mask}(\mathbf{G}, \mathbf{P}) = \mathcal{L}_{bce}(\mathbf{G}, \mathbf{P}) + \mathcal{L}_{dice}(\mathbf{G}, \mathbf{P}),
\end{equation}
where
\begin{equation}
\mathcal{L}_{bce}(\mathbf{G}, \mathbf{P}) = -\frac{1}{N} \sum_{j=1}^{N} [\mathbf{G}_j \log \mathbf{P}_j + (1-\mathbf{G}_j)\log(1-\mathbf{P}_j)],
\end{equation}
\begin{equation}
\mathcal{L}_{dice}(\mathbf{G}, \mathbf{P}) = 1 - \frac{2 \sum_{j=1}^{N} \mathbf{P}_j \mathbf{G}_j}{\sum_{j=1}^{N} \mathbf{P}_j + \sum_{j=1}^{N} \mathbf{G}_j}.
\end{equation}

To promote prediction consistency, a symmetric consistency loss is introduced to encourage similarity between the binarized predictions of $\mathbf{P}_a$ and $\mathbf{P}_b$. The binarization function $\mathcal{B}(\mathbf{P}, \theta)$ converts the continuous prediction map $\mathbf{P}$ into a binary mask according to a predefined threshold $\theta$:
\begin{equation}
\mathcal{B}(\mathbf{P}, \theta) =
\begin{cases}
1, & \text{if } \mathbf{P} \geq \theta, \\
0, & \text{otherwise.}
\end{cases}
\end{equation}
The consistency loss is then formulated as:
\begin{equation}
\mathcal{L}_{cons}(\mathbf{P}_a, \mathbf{P}_b) = \tfrac{1}{2}[\mathcal{L}_{bce}(\mathcal{B}(\mathbf{P}_b,\theta), \mathbf{P}_a) + \mathcal{L}_{bce}(\mathcal{B}(\mathbf{P}_a,\theta), \mathbf{P}_b)],
\end{equation}
which enforces the encoder to generate invariant predictions across input perturbations. The total loss function in this pretraining stage is defined as:
\begin{equation}
\mathcal{L}_{pretrain} = \mathcal{L}_{mask}(\mathbf{G}, \mathbf{P}_a) + \mathcal{L}_{mask}(\mathbf{G}, \mathbf{P}_b) + \mathcal{L}_{cons}(\mathbf{P}_a, \mathbf{P}_b).
\end{equation}
This objective enables the encoder to learn perturbation-invariant and semantically stable features, offering a robust initialization for subsequent distillation and cross-modal alignment.

\begin{figure}[htbp]
\centerline{\includegraphics[scale=0.31]{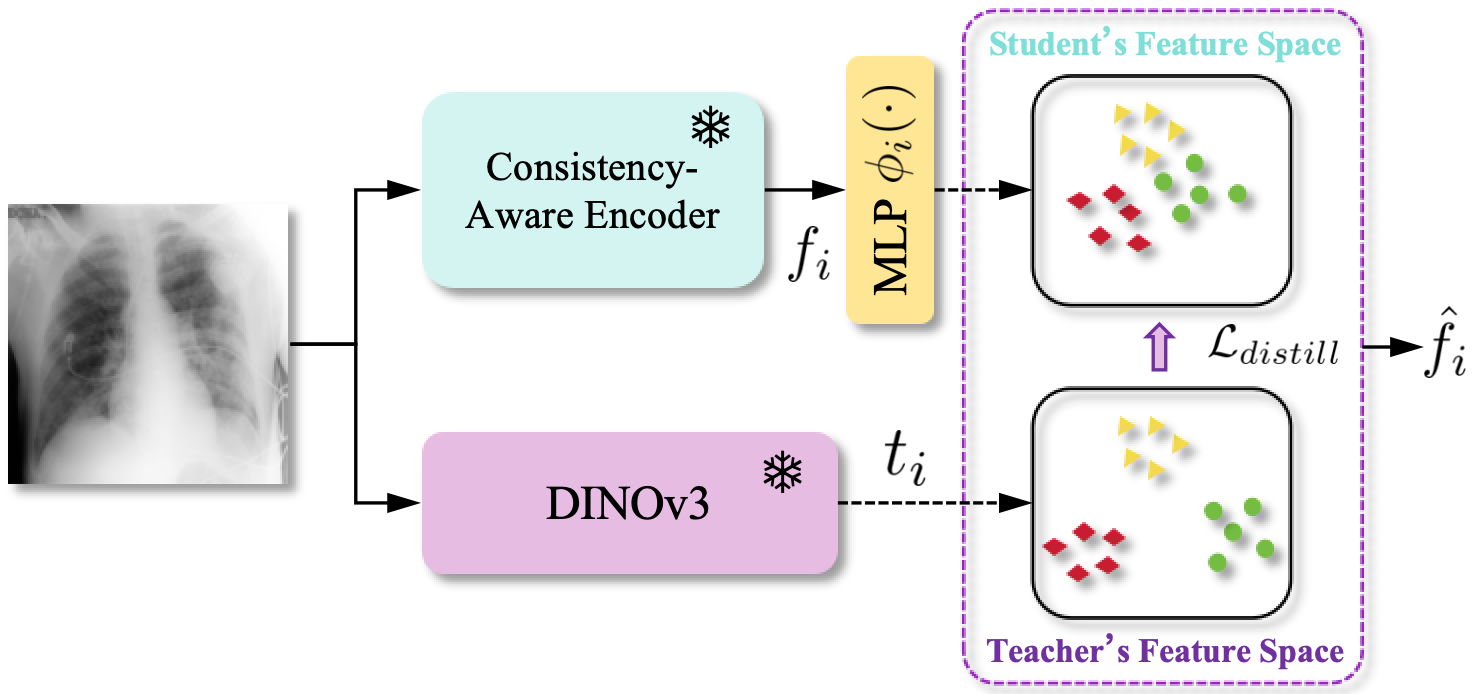}}
\caption{Overview of the semantic encoder distillation process, where features extracted by the frozen Consistency-Aware Encoder are refined under the supervision of the frozen DINOv3 teacher to obtain more discriminative and semantically enriched representations.}
\label{fig2}
\end{figure}

\begin{figure*}[htbp]
\centerline{\includegraphics[scale=0.35]{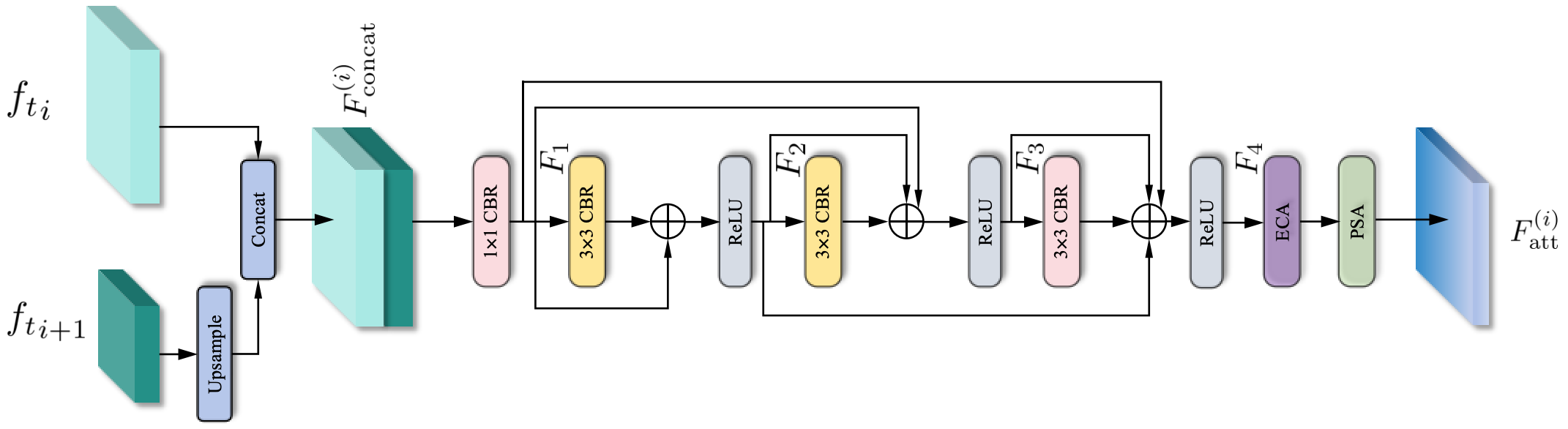}}
\caption{Architecture of the Scale-Adaptive Decoder (SAD). In this design, ECA and PSA denote the channel-attention and spatial-attention modules, respectively.}
\label{fig3}
\end{figure*}

\subsection{Semantic Encoder Distillation}
As illustrated in Fig~\ref{fig2}, to enhance the semantic discriminability of the encoder features, we introduce a semantic encoder distillation module guided by a frozen DINOv3~\cite{simeoni2025dinov3}, a powerful self-supervised visual encoder. This strategy substantially improves semantic correspondence with the pre-trained semantic features derived from the text encoder~\cite{Prism2025}. The first-stage encoder produces four hierarchical feature maps $f_i$ $(i=1,2,3,4)$, while the corresponding DINOv3 teacher features $t_i$ $(i=1,2,3,4)$ are obtained as normalized patch embeddings from its final transformer layers. Each $f_i$ is projected through a lightweight MLP $\phi_i(\cdot)$ to match the dimensionality of $t_i$. The distillation loss is defined using cosine similarity between normalized feature pairs:
\begin{equation}
\mathcal{L}_{distill} = -\frac{1}{N}\sum_{i=1}^{4}\sum_{j=1}^{L_i}
\frac{\phi_i(f_i^{(j)}) \cdot t_i^{(j)}}{\|\phi_i(f_i^{(j)})\|\|t_i^{(j)}\|},
\end{equation}
where $N=\sum_{i=1}^{4}L_i$ represents the total number of spatial tokens across all feature levels, and $L_i$ denotes the number of spatial tokens (i.e., patch embeddings) in the $i$-th feature level after flattening the spatial dimensions of $f_i$. This process encourages the encoder to learn semantically richer and more discriminative multi-scale representations, yielding distilled feature $\hat{f}_i$ ($i=1,2,3,4$) that provides a stronger foundation for subsequent cross-modal alignment with textual prompts.

\subsection{Cross-modal Alignment}

To incorporate semantic cues from clinical text prompts into visual representations, we design a Cross-modal Alignment Module that enables fine-grained interaction between textual embeddings and multi-scale image features. Specifically, the hierarchical encoder features $\hat{f}_i$ ($i=1,2,3,4$) are aligned with the text embedding $\mathbf{T}$ extracted from a frozen CXR-BERT model~\cite{boecking2022making}. Following standard cross-modal attention mechanisms, the visual features serve as queries, while the text embeddings act as keys and values, allowing language semantics to guide visual feature refinement. Formally, for a visual feature map $\hat{f}_i \in \mathbb{R}^{H_i \times W_i \times C}$ and text embedding $\mathbf{T} \in \mathbb{R}^{L \times C}$, the cross-attention is computed as:
\begin{equation}
{f_t}_i = \operatorname{Softmax} \left( (\hat{f}_i W_q)(\mathbf{T} W_k)^T / \sqrt{d} \right)(\mathbf{T} W_v),
\end{equation}
where $W_q, W_k, W_v$ are learnable projection matrices, and $d$ is a scaling factor. The output ${f_t}_i$ represents the text-enhanced visual features at scale $i$.

\begin{table*}[t]
\centering
\caption{Comparison with state-of-the-art uni-modal and multi-modal segmentation methods on three benchmark datasets using Dice (\%) and mIoU (\%). Best results are highlighted in \textbf{bold}. The \textbf{Text} column indicates whether a method is uni-modal (\ding{55}) or multi-modal (\ding{51}). 
} 
\resizebox{1.90\columnwidth}{!}{
\begin{tabular}{l|c|c|c|c|cc|cc|cc}
\toprule
\textbf{Method} & \textbf{Backbone} &\textbf{Params} ↓ (M) & \textbf{FLOPs} ↓ (G) & \textbf{Text} & \multicolumn{2}{c|}{\textbf{Kvasir-SEG}} & \multicolumn{2}{c|}{\textbf{MosMedData+}} & \multicolumn{2}{c}{\textbf{QaTa-COV19}} \\
 &  & & & & Dice ↑ & mIoU ↑ & Dice ↑ & mIoU ↑ & Dice ↑ & mIoU ↑ \\
\midrule
U-Net~\cite{unet} & CNN & 31.04 & 54.75 & \ding{55} & 78.94 & 70.21 & 64.60 & 50.73 & 79.02 & 69.46 \\
U-Net++~\cite{zhou2018unet++} & CNN & 36.01 & 25.22 &\ding{55} & 80.35 & 71.85 & 71.75 & 58.39 & 79.62 & 70.25 \\
nnU-Net~\cite{isensee2021nnu} & CNN & 41.25& 32.71 & \ding{55} & 81.42 & 73.10 & 72.59 & 60.36 & 80.42 & 70.81 \\
TransUNet~\cite{transunet} & Hybrid & 104.93 & 37.22 & \ding{55} & 82.05 & 74.12 & 71.24 & 58.44 & 78.63 & 69.13 \\
Swin-UNet~\cite{cao2022swin} & Hybrid &62.03& 42.37& \ding{55} & 82.47 & 74.35 & 63.29 & 50.19 & 78.07 & 68.34 \\
MAXFormer~\cite{liang2023maxformer} & Hybrid &68.96 &44.87 &\ding{55} & 83.02 & 75.26 & 65.90 & 52.69 & 79.15 & 69.60 \\
ConDSeg~\cite{lei2025condseg} & CNN & 43.81 & 28.62 & \ding{55} & 89.89 & 82.73 & 75.89 & 65.18 & 89.76 & 78.38 \\
MedSegDiff-V2~\cite{wu2024medsegdiff} & Diffusion & 46.12 & 149.01 & \ding{55} & 90.97 & 83.27 & 78.59 & 64.55 & 90.81 & 83.04 \\
\midrule
TextDiff~\cite{feng2024enhancing} & Diffusion & 9.68 & 36.85 & \ding{51} & 83.43 & 76.03 & 76.52 & 62.77 & 80.86 & 77.91\\
LViT~\cite{li2023lvit} & Hybrid & 93.16& 46.75&\ding{51} & 86.54 & 78.33 & 74.57 & 61.33 & 83.66 & 75.11 \\
LGA~\cite{hu2024lga} & Transformer &97.58 & 48.23 &\ding{51} & 87.26 & 79.42 & 75.63 & 62.52 & 84.65 & 76.23 \\
CausalCLIPSeg~\cite{chen2024causalclipseg} & Hybrid &89.86 &45.95 &\ding{51} & 87.89 & 79.95 & 76.12 & 63.05 & 85.21 & 76.90 \\
TVE-Net~\cite{fang2025driven} & CNN &56.26  & 33.46 & \ding{51} & 88.04 & 80.11 & 77.48 & 65.07 & 85.22 & 77.00 \\
TGCAM~\cite{guo2024common} & CNN &61.76& 35.99& \ding{51} & 88.85 & 81.03 & 77.73 & 65.32 & 87.41 & 77.82 \\
LanGuideSeg~\cite{zhong2023ariadne} & CNN &64.35&36.27 & \ding{51} & 89.93 & 82.12 & 78.11 & 66.05 & 89.78 & 81.45 \\
MAdapter~\cite{zhang2024madapter} & CNN & 67.55 & 37.65& \ding{51} & 90.16 & 82.65 & 78.62 & 64.78 & 90.22 & 82.16 \\
ProLearn~\cite{ye2025alleviating} & CNN &69.88 & 38.12 & \ding{51} & 90.40 & 82.80 & 77.82 & 63.69 & 90.60 & 82.81 \\
\textbf{TAMISeg (ours)} & CNN &48.75& 29.83 & \ding{51} & \textbf{91.59} & \textbf{84.11} & \textbf{79.30} & \textbf{65.71} & \textbf{91.79
} & \textbf{84.95} \\
\bottomrule
\end{tabular}
}
\label{tab:comparison} 
\end{table*}

\begin{figure*}[t]
\centering
\setlength{\abovecaptionskip}{2pt}
\setlength{\belowcaptionskip}{-4pt}

\includegraphics[width=0.67\textwidth]{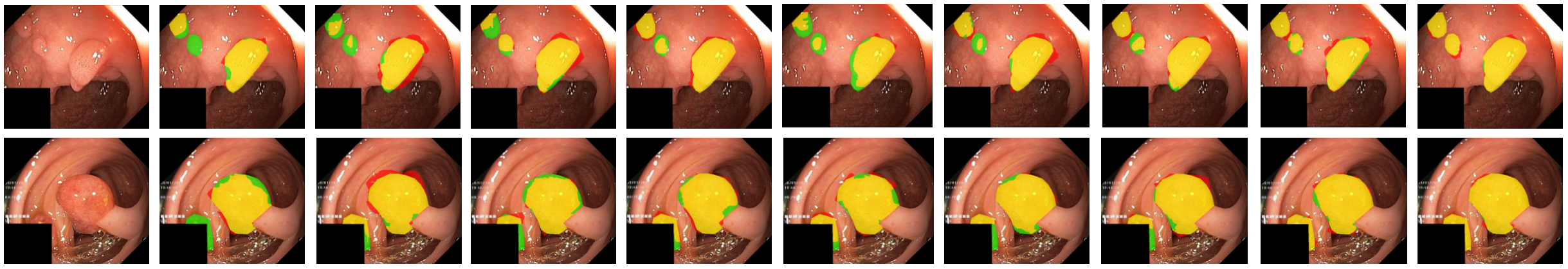}
\vspace{2pt}
\includegraphics[width=0.67\textwidth]{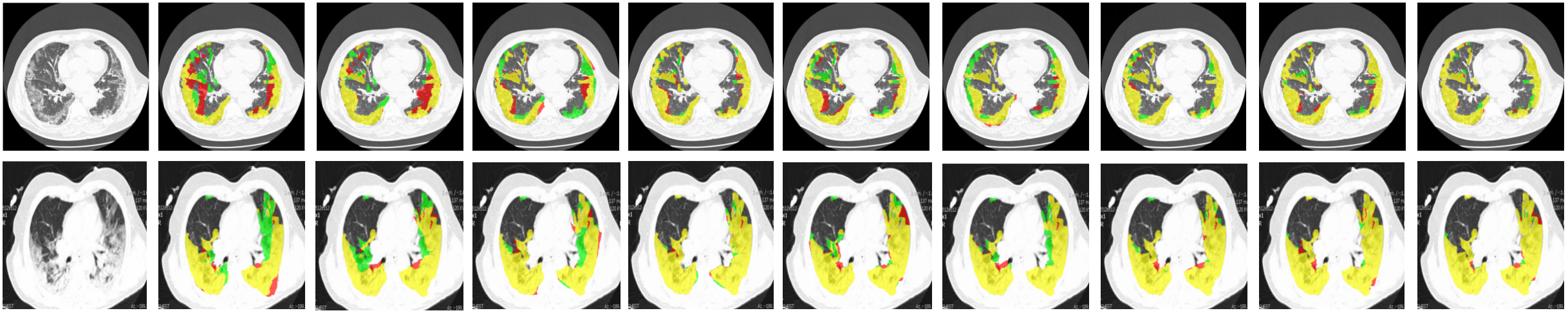}
\vspace{1pt}
\includegraphics[width=0.67\textwidth]{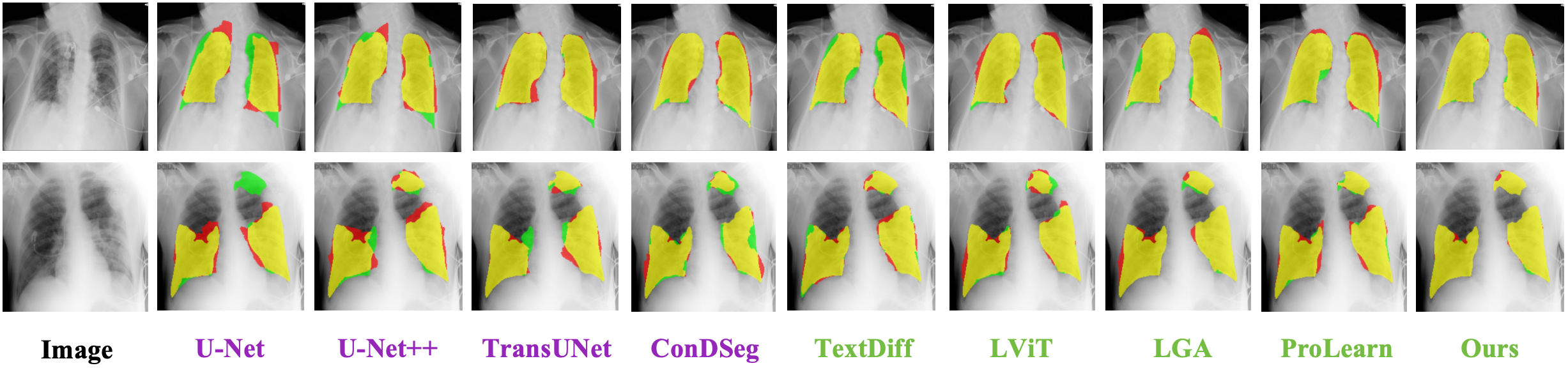}

\caption{Qualitative comparison of segmentation results on three benchmark datasets. 
From top to bottom: Kvasir-SEG, MosMedData+, and QaTa-COV19. 
Yellow regions denote overlap between prediction and ground truth, while red and green areas indicate over- and under-segmentation, respectively. 
TAMISeg achieves more accurate and consistent delineations, with larger yellow overlap regions and clearer boundaries across varying lesion sizes and contrast conditions. Model names are color-coded: purple indicates uni-modal methods, and green represents multi-modal methods.}
\label{fig:qualitative_results}
\end{figure*}


\subsection{Scale-Adaptive Decoder}
As shown in Fig~\ref{fig3}, to handle anatomical structures of varying sizes, we adopt a Scale-Adaptive Decoder consisting of three parallel branches that process text-aligned multi-scale features, each specialized for small, medium, and large anatomical structures~\cite{lei2025condseg}. Each decoder branch takes a pair of adjacent feature maps \(({f_t}_i, {f_t}_{i+1})\) obtained from the encoder after cross-modal alignment. The shallower feature ${f_t}_i$ preserves fine-grained spatial details, whereas the deeper feature ${f_t}_{i+1}$ provides broader contextual and semantic information for recognizing larger structures. To combine them, the lower-resolution ${f_t}_{i+1}$ is first upsampled via bilinear interpolation and then concatenated with ${f_t}_i$ along the channel dimension:
\begin{equation}
F_{\text{concat}}^{(i)} = \text{Concat}(\text{Up}({f_t}_{i+1}), {f_t}_i),
\end{equation}
where \(\text{Up}(\cdot)\) denotes $2\times$ upsampling. The merged feature \(F_{\text{concat}}^{(i)}\) is refined using a series of residual convolutional operations:
\begin{equation}
\begin{split}
F_1 &= C_1(F_{\text{concat}}^{(i)}), \\
F_2 &= \text{ReLU}(C_2(F_1) + F_1), \\
F_3 &= \text{ReLU}(C_3(F_2) + F_2 + F_1), \\
F_4 &= \text{ReLU}(C_4(F_3) + F_3 + F_2 + F_1),
\end{split}
\end{equation}
where $C_1$ and $C_4$ denote $1 \times 1$ convolutions, and $C_2$, $C_3$ are $3 \times 3$ convolutions. Residual connections promote gradient flow and preserve hierarchical cues. An Efficient Channel Attention (ECA)~\cite{wang2020eca} and a Pyramid Spatial Attention (PSA)~\cite{zhao2018psanet} module are sequentially applied to enhance salient features while suppressing redundant ones:
\begin{equation}
F_{\text{att}}^{(i)} = \text{PSA}(\text{ECA}(F_4)).
\end{equation}

This process is repeated across three decoder branches—Decoder\(_s\), Decoder\(_m\), and Decoder\(_l\)—receiving feature pairs \(( {f_t}_1, {f_t}_2 )\), \(( {f_t}_2, {f_t}_3 )\), and \(( {f_t}_3, {f_t}_4 )\), respectively. The resulting outputs are upsampled to a uniform resolution and concatenated:
\begin{equation}
F_{\text{final}} = \operatorname{Concat}\big(F_{\text{att}}^{(s)},\ \operatorname{Up}_2(F_{\text{att}}^{(m)}),\ \operatorname{Up}_4(F_{\text{att}}^{(l)})\big),
\end{equation}
where \(\text{Up}_2(\cdot)\) and \(\text{Up}_4(\cdot)\) denote $2\times$ and $4\times$ upsampling. The fused representation is finally processed by a lightweight convolutional head to generate the predicted segmentation mask:
\begin{equation}
Mask_{pred} = \sigma(C_6(C_5(F_{\text{final}}))),
\end{equation}
where $C_5$ and $C_6$ are $3\times3$ and $1\times1$ convolutions, and $\sigma$ is the Sigmoid activation. The SAD thus allows TAMISeg to perform fine-grained, scale-adaptive segmentation by jointly leveraging detailed local and global contextual information.

The overall training objective of TAMISeg combines the segmentation prediction loss with the semantic encoder distillation loss to jointly optimize segmentation accuracy. The final loss function is defined as:
\begin{equation}
\mathcal{L}_{\text{total}} = \mathcal{L}_{\text{pred}} + \lambda \mathcal{L}_{\text{distill}},
\end{equation}
where $\lambda$ balances the contribution of the distillation loss during optimization.

\section{Experiments}
\subsection{Experimental Setup}
All experiments were conducted on a single NVIDIA GeForce RTX 3090 GPU with 24 GB of memory. Images were resized to $256 \times 256$ pixels for both training and evaluation. The training consisted of two stages: the encoder was first pretrained independently using the Adam optimizer with a learning rate of $1\times10^{-4}$, a batch size of 8, and 300 epochs with early stopping (patience = 100); then the full TAMISeg framework was fine-tuned end-to-end with pretrained encoder weights, a batch size of 4, and a learning rate of $1\times10^{-5}$ for 100 epochs using bilinear interpolation for upsampling.

\subsection{Comparison of Segmentation Performance}
To comprehensively evaluate the performance of the proposed TAMISeg framework, we compare it with a series of state-of-the-art segmentation methods, including both uni-modal and multi-modal approaches. The uni-modal baselines include several classic models, namely U-Net~\cite{unet}, U-Net++~\cite{zhou2018unet++}, nnU-Net~\cite{isensee2021nnu}, TransUNet~\cite{transunet}, Swin-UNet~\cite{cao2022swin}, MAXFormer~\cite{liang2023maxformer}, and ConDSeg~\cite{lei2025condseg}, which represent mainstream CNN-based and hybrid CNN–Transformer segmentation architectures. For multi-modal methods, we include representative text-guided segmentation frameworks such as LViT~\cite{li2023lvit}, TextDiff~\cite{feng2024enhancing}, LGA~\cite{hu2024lga}, CausalCLIPSeg~\cite{chen2024causalclipseg}, TVE-Net~\cite{fang2025driven}, LanGuideSeg~\cite{zhong2023ariadne}, MAdapter~\cite{zhang2024madapter}, TGCAM~\cite{guo2024common}, and ProLearn~\cite{ye2025alleviating}, all of which incorporate clinical linguistic information for semantic enhancement.

All models are trained and evaluated under identical experimental settings for a fair comparison. Dice coefficient and mean Intersection-over-Union (mIoU) are adopted as the primary quantitative evaluation metrics. As summarized in Table~\ref{tab:comparison}, TAMISeg consistently outperforms both uni-modal and multi-modal baselines across three benchmark datasets: Kvasir-SEG, MosMedData+, and QaTa-COV19. These results confirm that the integration of consistency-aware encoder, DINOv3-based semantic encoder distillation, and scale-adaptive decoding significantly enhances both robustness and segmentation accuracy, particularly in complex or low-contrast imaging conditions. Qualitative visualizations in Fig.~\ref{fig:qualitative_results} further support these findings, where TAMISeg demonstrates more precise and complete segmentations—especially for lesions of varying sizes, as observed in the first and last row examples.

\begin{table}[t]
\centering
\caption{Ablation study on the Kvasir-SEG dataset. Modules are added sequentially to evaluate their individual contributions.}
\label{tab:ablation}
\setlength{\tabcolsep}{3pt}
\renewcommand{\arraystretch}{1.05}
\begin{tabular}{lccc|cc}
\toprule
\textbf{Variant} & CMA & SED & SAD & Dice (\%) & mIoU (\%) \\
\midrule
CAE (Baseline)     & \ding{55} & \ding{55} & \ding{55} & 88.42 & 81.03 \\
+ CMA              & \ding{51} & \ding{55} & \ding{55} & 89.26 & 82.04 \\
+ CMA + SED        & \ding{51} & \ding{51} & \ding{55} & 90.28 & 83.12 \\
Full (CMA+SED+SAD) & \ding{51} & \ding{51} & \ding{51} & \textbf{91.59} & \textbf{84.11} \\
\bottomrule
\end{tabular}
\end{table}

\subsection{Ablation Studies}
To explore the contribution of each component in TAMISeg, we performed ablation experiments on the Kvasir-SEG dataset. Starting from the \textit{Consistency-Aware Encoder} (CAE) as the baseline, we progressively incorporated the \textit{Cross-modal Alignment} (CMA), \textit{Semantic Encoder Distillation } (SED), and \textit{Scale-Adaptive Decoder} (SAD). As shown in Table~\ref{tab:ablation}, each additional component yields consistent performance improvements in both Dice and mIoU metrics. The CMA module effectively integrates semantic cues from clinical text, the SED enhances the discriminative power of encoder features through distillation, and the SAD improves segmentation precision across different anatomical scales. The complete TAMISeg achieves the best overall performance, demonstrating the complementary benefits of all modules.

\subsection{Conclusions}
In this paper, we propose TAMISeg, a text-aligned medical image segmentation framework that leverages semantic encoder distillation to complement pixel-level consistency-aware encoding. The pretrained consistency-aware encoder improves stability under pixel-level, while the distillation process further enriches the semantic representations along with the pre-trained text encoder. A scale-adaptive decoder enables accurate segmentation across different anatomical scales. Experiments on three public datasets show that TAMISeg outperforms both uni-modal and multi-modal methods, highlighting its stability and adaptability to diverse anatomical structures and imaging conditions.

\bibliographystyle{IEEEbib}
\bibliography{icme2025references_short}

@article{Prism2025,
  title={The Prism Hypothesis: Harmonizing Semantic and Pixel Representations via Unified Autoencoding},
  author={Fan, Weichen and Diao, Haiwen and Wang, Quan and Lin, Dahua and Liu, Ziwei},
  journal={arXiv preprint arXiv:2512.19693},
  year={2025}
}

@inproceedings{unet,
  title     = {{U-Net}: Convolutional Networks for Biomedical Image Segmentation},
  author    = {Ronneberger, Olaf and Fischer, Philipp and Brox, Thomas},
  booktitle = {MICCAI},
  pages     = {234--241},
  year      = {2015}
}

@article{transunet,
  title={Transunet: Transformers make strong encoders for medical image segmentation},
  author={Chen, Jieneng and Lu, Yongyi and Yu, Qihang and Luo, Xiangde and Adeli, Ehsan and Wang, Yan and Lu, Le and Yuille, Alan L and Zhou, Yuyin},
  journal={arXiv preprint arXiv:2102.04306},
  year={2021}
}

@inproceedings{zhou2018unet++,
  title     = {{U-Net++}: A Nested U-Net Architecture for Medical Image Segmentation},
  author    = {Zhou, Zongwei and Rahman Siddiquee, Md Mahfuzur and Tajbakhsh, Nima and Liang, Jianming},
  booktitle = {DLMIA},
  pages     = {3--11},
  year      = {2018}
}

@article{isensee2021nnu,
  title={nnU-Net: a self-configuring method for deep learning-based biomedical image segmentation},
  author={Isensee, Fabian and Jaeger, Paul F and Kohl, Simon AA and Petersen, Jens and Maier-Hein, Klaus H},
  journal={Nature methods},
  volume={18},
  number={2},
  pages={203--211},
  year={2021},
  publisher={Nature Publishing Group}
}

@article{liang2023maxformer,
  title     = {MAXFormer: Enhanced Transformer for Medical Image Segmentation with Multi-Attention and Multi-Scale Feature Fusion},
  author    = {Liang, Zhiwei and Zhao, Kui and Liang, Gang and Li, Siyu and Wu, Yifei and Zhou, Yiping},
  journal   = {Knowl.-Based Syst.},
  volume    = {280},
  pages     = {110987},
  year      = {2023}
}

@article{fang2025driven,
  title     = {Driven by Textual Knowledge: A Text-View Enhanced Knowledge Transfer Network for Lung Infection Region Segmentation},
  author    = {Fang, Lexin and Li, Xuemei and Xu, Yunyang and Zhang, Fan and Zhang, Caiming},
  journal   = {Med. Image Anal.},
  pages     = {103625},
  year      = {2025}
}

@inproceedings{zhong2023ariadne,
  title     = {Ariadne’s Thread: Using Text Prompts to Improve Segmentation of Infected Areas from Chest X-ray Images},
  author    = {Zhong, Yi and Xu, Mengqiu and Liang, Kongming and Chen, Kaixin and Wu, Ming},
  booktitle = {MICCAI},
  pages     = {724--733},
  year      = {2023}
}

@inproceedings{chen2024causalclipseg,
  title     = {{CausalCLIPSeg}: Unlocking CLIP’s Potential in Referring Medical Image Segmentation with Causal Intervention},
  author    = {Chen, Yaxiong and Wei, Minghong and Zheng, Zixuan and Hu, Jingliang and Shi, Yilei and Xiong, Shengwu and Zhu, Xiao Xiang and Mou, Lichao},
  booktitle = {MICCAI},
  pages     = {77--87},
  year      = {2024}
}

@inproceedings{zhang2024madapter,
  title     = {{MAdapter}: A Better Interaction Between Image and Language for Medical Image Segmentation},
  author    = {Zhang, Xu and Ni, Bo and Yang, Yang and Zhang, Lefei},
  booktitle = {MICCAI},
  pages     = {425--434},
  year      = {2024}
}

@inproceedings{guo2024common,
  title     = {Common Vision-Language Attention for Text-Guided Medical Image Segmentation of Pneumonia},
  author    = {Guo, Yunpeng and Zeng, Xinyi and Zeng, Pinxian and Fei, Yuchen and Wen, Lu and Zhou, Jiliu and Wang, Yan},
  booktitle = {MICCAI},
  pages     = {192--201},
  year      = {2024}
}

@inproceedings{feng2024enhancing,
  title     = {Enhancing Label-efficient Medical Image Segmentation with Text-guided Diffusion Models},
  author    = {Feng, Chun-Mei},
  booktitle = {MICCAI},
  pages     = {253--262},
  year      = {2024}
}

@inproceedings{hu2024lga,
  title     = {{LGA}: A Language Guide Adapter for Advancing the SAM Model’s Capabilities in Medical Image Segmentation},
  author    = {Hu, Jihong and Li, Yinhao and Sun, Hao and Song, Yu and Zhang, Chujie and Lin, Lanfen and Chen, Yen-Wei},
  booktitle = {MICCAI},
  pages     = {610--620},
  year      = {2024}
}

@article{li2023lvit,
  title     = {{LViT}: Language Meets Vision Transformer in Medical Image Segmentation},
  author    = {Li, Zihan and Li, Yunxiang and Li, Qingde and Wang, Puyang and Guo, Dazhou and Lu, Le and Jin, Dakai and Zhang, You and Hong, Qingqi},
  journal   = {IEEE Trans. Med. Imaging},
  volume    = {43},
  number    = {1},
  pages     = {96--107},
  year      = {2023}
}

@inproceedings{ye2025alleviating,
  title     = {Alleviating Textual Reliance in Medical Language-guided Segmentation via Prototype-driven Semantic Approximation},
  author    = {Ye, Shuchang and Naseem, Usman and Meng, Mingyuan and Kim, Jinman},
  booktitle = {ICCV},
  pages     = {22316--22326},
  year      = {2025}
}

@inproceedings{lei2025condseg,
  title     = {{ConDSeg}: A General Medical Image Segmentation Framework via Contrast-Driven Feature Enhancement},
  author    = {Lei, Mengqi and Wu, Haochen and Lv, Xinhua and Wang, Xin},
  booktitle = {AAAI},
  pages     = {4571--4579},
  year      = {2025}
}

@article{li2025kg,
  title={{KG-SAM}: Injecting Anatomical Knowledge into Segment Anything Models via Conditional Random Fields},
  author={Li, Yu and Chang, Da and Xiao, Xi},
  journal={arXiv preprint arXiv:2509.21750},
  year={2025}
}

@article{jie2025diffusion,
  title={Diffusion-Guided Mask-Consistent Paired Mixing for Endoscopic Image Segmentation},
  author={Jie, Pengyu and Liu, Wanquan and He, Rui and Wen, Yihui and Meng, Deyu and Gao, Chenqiang},
  journal={arXiv preprint arXiv:2511.03219},
  year={2025}
}

@inproceedings{wang2022medclip,
  title     = {{MedCLIP}: Contrastive Learning from Unpaired Medical Images and Text},
  author    = {Wang, Zifeng and Wu, Zhenbang and Agarwal, Dinesh and Sun, Jimeng},
  booktitle = {EMNLP},
  pages     = {3876--3886},
  year      = {2022}
}

@article{simeoni2025dinov3,
  title={{DINOv3}},
  author={Sim{\'e}oni, Oriane and Vo, Huy V and Seitzer, Maximilian and Baldassarre, Federico and Oquab, Maxime and Jose, Cijo and Khalidov, Vasil and Szafraniec, Marc and Yi, Seungeun and Ramamonjisoa, Micha{\"e}l and others},
  journal={arXiv preprint arXiv:2508.10104},
  year={2025}
}

@inproceedings{cao2022swin,
  title     = {{Swin-Unet}: Unet-like Pure Transformer for Medical Image Segmentation},
  author    = {Cao, Hu and Wang, Yueyue and Chen, Joy and Jiang, Dongsheng and Zhang, Xiaopeng and Tian, Qi and Wang, Manning},
  booktitle = {ECCV},
  pages     = {205--218},
  year      = {2022}
}

@inproceedings{boecking2022making,
  title     = {Making the Most of Text Semantics to Improve Biomedical Vision--Language Processing},
  author    = {Boecking, Benedikt and Usuyama, Naoto and Bannur, Shruthi and Castro, Daniel C. and Schwaighofer, Anton and Hyland, Stephanie and Wetscherek, Maria and Naumann, Tristan and Nori, Aditya and Alvarez-Valle, Javier and others},
  booktitle = {ECCV},
  pages     = {1--21},
  year      = {2022}
}

@inproceedings{wang2020eca,
  title     = {{ECA-Net}: Efficient Channel Attention for Deep Convolutional Neural Networks},
  author    = {Wang, Qilong and Wu, Banggu and Zhu, Pengfei and Li, Peihua and Zuo, Wangmeng and Hu, Qinghua},
  booktitle = {CVPR},
  pages     = {11534--11542},
  year      = {2020}
}

@inproceedings{zhao2018psanet,
  title     = {{PSANet}: Point-Wise Spatial Attention Network for Scene Parsing},
  author    = {Zhao, Hengshuang and Zhang, Yi and Liu, Shu and Shi, Jianping and Loy, Chen Change and Lin, Dahua and Jia, Jiaya},
  booktitle = {ECCV},
  pages     = {267--283},
  year      = {2018}
}

@inproceedings{wu2024medsegdiff,
  title     = {{MedSegDiff-V2}: Diffusion-based Medical Image Segmentation with Transformer},
  author    = {Wu, Junde and Ji, Wei and Fu, Huazhu and Xu, Min and Jin, Yueming and Xu, Yanwu},
  booktitle = {AAAI},
  pages     = {6030--6038},
  year      = {2024}
}

@article{yan2024cold,
  title   = {{Cold SegDiffusion}: A Novel Diffusion Model for Medical Image Segmentation},
  author  = {Yan, Pengfei and Li, Minglei and Zhang, Jiusi and Li, Guanyi and Jiang, Yuchen and Luo, Hao},
  journal = {Knowl.-Based Syst.},
  volume  = {301},
  pages   = {112350},
  year    = {2024}
}



\end{document}